\documentclass[akbc,twoside,11pt,lettersize]{article}
\usepackage{akbc}
\akbcheading
\ShortHeadings{Training Vision-Language Models with Less Bimodal Supervision}{Segal, Bogin \& Berant}
\finalcopy % Uncomment for camera-ready version, but NOT for submission.

\usepackage{diagbox}
\usepackage{amsfonts} % \mathbb
\usepackage{amsmath} % \begin{align*}
\usepackage{bm} % \bm
\usepackage[capitalise]{cleveref} % \cref
\usepackage{multirow}
\usepackage{graphicx}
\usepackage{makecell}
\graphicspath{ {./images/} }
\usepackage{textcomp} % for PyTorch citation
\usepackage{booktabs} % nice tables
\usepackage{makecell}
\usepackage{comment}
\usepackage{wasysym}
\usepackage{floatrow}
\usepackage{wrapfig}
\usepackage[shortlabels]{enumitem}

\newfloatcommand{capbtabbox}{table}[][\FBwidth]

\newcommand\sC{\mathcal{C}}

\newcommand\bmx{\bm{x}}
\newcommand\bmi{\bm{i}}
\newcommand\commentout[1]{}

\begin{document}

\title{Training Vision-Language Models with Less Bimodal Supervision}

\author{\name Elad Segal \email elad.segal@gmail.com \\ 
\name Ben Bogin \email ben.bogin@cs.tau.ac.il \\
\name Jonathan Berant \email joberant@cs.tau.ac.il \\
       \addr Blavatnik School of Computer Science\\
       Tel Aviv University}

% For research notes, remove the comment character in the line below.
% \researchnote

\maketitle

\begin{abstract}

Standard practice in pretraining multimodal models, such as vision-language models, is to rely on pairs of aligned inputs from both modalities, for example, aligned image-text pairs.
However, such pairs can be difficult to obtain in low-resource settings and for some modality pairs (e.g., structured tables and images).
In this work, we investigate the extent to which we can reduce the reliance on such parallel data, which we term \emph{bimodal supervision}, and use models that are pretrained on each modality independently. We experiment with a high-performing vision-language model, and analyze the effect of bimodal supervision on three vision-language tasks. We find that on simpler tasks, such as VQAv2 and GQA, one can eliminate bimodal supervision completely, suffering only a minor loss in performance. Conversely, for NLVR2, which requires more complex reasoning, training without bimodal supervision leads to random performance. Nevertheless, using only 5\% of the bimodal data (142K images along with their captions), or leveraging weak supervision in the form of a list of machine-generated labels for each image, leads to only a moderate degradation compared to using 3M image-text pairs: 74\%$\rightarrow$$\sim$70\%.

\end{abstract}
\section{Introduction}

\begin{figure}[ht]
   \centering
   \includegraphics[scale=0.365]{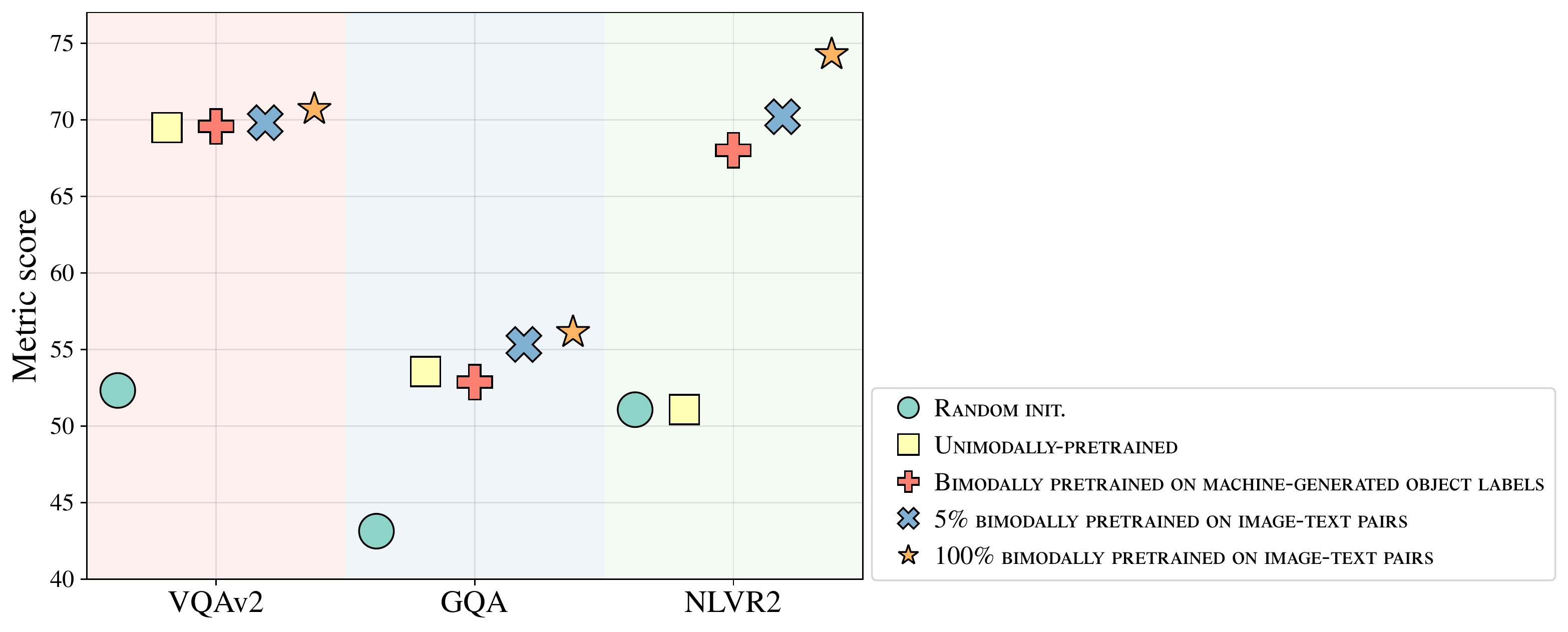}
 \vspace{-0.4cm}
  \caption{
  The effect of unimodal and bimodal pretraining on downstream performance after finetuning. In VQAv2 and GQA, pretraining on unimodal data alone (without image-text pairs) is competitive with models pretrained on image-text pairs. On NLVR2, bimodal supervision is necessary, but one can reach reasonable performance using only 5\% of the image-text pairs or training on machine-generated object labels. Random initialization leads to poor performance on all tasks.} 
  \label{fig:main_results}
\end{figure}

Pretraining models on large amounts of raw data using self-supervision has revolutionized machine learning, and is now standard practice across a wide range of modalities \cite{Liu2019RoBERTaAR,JMLR:v21:20-074,DosovitskiyB0WZ21,Liu_2021_ICCV,herzig-etal-2020-tapas,schneider2019wav2vec,baevski2022data2vec}.
While typically pretrained models are trained on data from a single modality (\emph{unimodal data}), the success of pretraining has spread to the \emph{bimodal} setup, where models are trained on pairs of inputs, each from a different modality (e.g. text and audio, \citealp{li-etal-2021-ctal}). Most notably, vision-language models, such as LXMERT \cite{tan-bansal-2019-lxmert}, ViLT \cite{pmlr-v139-kim21k}, METER \cite{Dou_2022_CVPR}, CLIP \cite{radford2021learning}, and others \cite{li2019visualbert,lu2019vilbert,ALBEF}, have been pretrained on manually or automatically collected parallel data that consists of aligned image-text pairs.

While effective, pretraining on bimodal data comes at a cost. First, gathering high-quality pairs can be challenging, especially in low-resource languages and domains, or for modality pairs where parallel data is scarce. Second, expanding this approach to more than two modalities (as in, e.g., MultimodalQA, \citealp{talmor2021multimodalqa}) is challenging. Last, pretraining is computationally expensive \citep{strubell-etal-2019-energy, Bommasani2021OnTO}, and thus relying on pretraining for all modality pairs is inefficient.

Given these shortcomings, a natural question is how far can we get with models pretrained on unimodal data only (\emph{unimodal models}), such as BERT \cite{devlin-etal-2019-bert} and ViT \citep{DosovitskiyB0WZ21}, to reduce or obviate the need for \emph{bimodal} pretraining. Can we align unimodal representations without resorting to pretraining over millions of input pairs?
While past work \cite{Dou_2022_CVPR,ALBEF,Zhai_2022_CVPR} used unimodal models as an initialization point before bimodal pretraining, it did not investigate its effect on the amount of necessary bimodal data. 

%What do we do
In this work, we investigate to what extent we can reduce the burden of bimodal pretraining and finetune models on vision-language applications starting with models that were unimodally pretrained. We choose a high-performing architecture \cite{Dou_2022_CVPR} -- a transformer image encoder and a transformer text encoder, which pass their representations through additional transformer layers that capture the interaction between the image and the text, before performing a final classification task.

We test performance on visual question answering and visual reasoning tasks in the following setups: (a) randomly initialized image and text encoders, (b) unimodally-pretrained image and text encoders, and (c) unimodally-pretrained image and text encoders that are then pretrained with bimodal supervision. We test different sources for bimodal pretraining,
which require different amounts of human effort: 
(a) automatically harvested image-caption pairs (Conceptual Captions, \citealp{sharma2018conceptual}),
(b) images paired with machine-generated object labels (CCIL, \citealp{ng-etal-2021-understanding}),
(c) manually annotated image-object pairs (ImageNet-1K, \citealp{ILSVRC15}),
and (d) image-question-answer triples from visual question answering tasks. We note that due to computational constraints the size of our pretraining corpus is smaller compared to those used by industry-based researchers \cite{li2022blip,radford2021learning,Jia2021ScalingUV}.

We find (Figure~\ref{fig:main_results}) that on some tasks, models that do not use any bimodal supervision are only slightly worse than models that are pretrained on large amounts of image-text pairs -- 70.7$\rightarrow$69.5 on VQAv2, and 56.1$\rightarrow$53.6 on GQA. However, for a more complex reasoning task, such as NLVR2, bimodal supervision is \emph{crucial}. Nevertheless, we show that one can dramatically reduce the number of bimodal image-text pairs and still obtain reasonable performance -- either by using only 5\% of the pairs (74.3$\rightarrow$70.2) or through machine-generated object labels (74.3$\rightarrow$68.0).
Our code is available at \href{https://github.com/eladsegal/less-bimodal-sup}{https://github.com/eladsegal/less-bimodal-sup}.

\section{Overview}

\begin{figure*}[ht]
   \centering
   \includegraphics[scale=0.495]{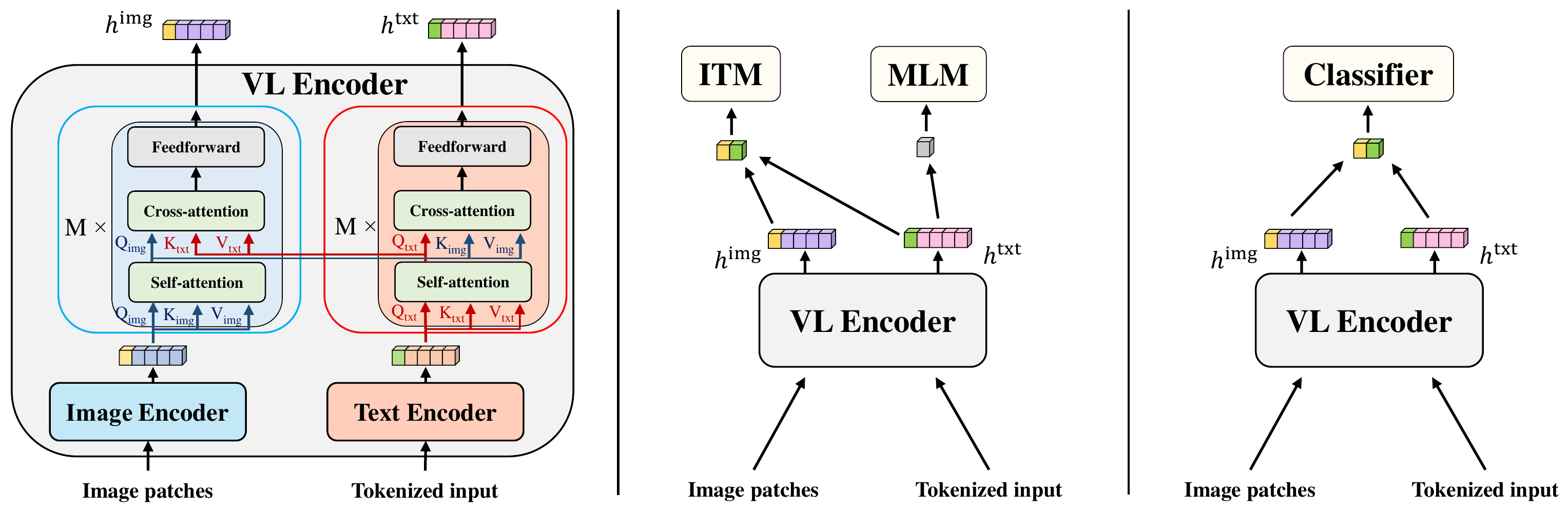}
   \vspace{-0.7cm}
  \caption{\emph{Left}: Architecture overview: an image encoder and a text encoder followed by a few transformer fusion layers, capturing interaction between modalities through cross-attention. \emph{Center}: We pretrain the VL encoder from bimodal supervision by taking contextualized representations of the image ($h^{\text{img}}$) and text ($h^{\text{txt}}$) and applying the image-text matching (ITM) and masked language modeling (MLM) loss functions. \emph{Right}: We finetune the VL encoder on downstream classification tasks by concatenating the image and text representations and passing them through an MLP classifier.} 
  \label{fig:overview}
\end{figure*}

We provide an overview of the experimental settings explored in this work. 
As our architecture-of-choice, we leverage one that has been shown to perform well across multiple tasks \cite{Dou_2022_CVPR}, namely, a Vision-Language (VL) encoder, where a unimodal image encoder creates image representations, a unimodal text encoder creates text representations, and these two representations are passed through a few transformer \citep{vaswani} layers that capture cross-modal interactions (Figure~\ref{fig:overview}, Left).

We experiment with three initializations of the image and text encoders. First, we use random initialization as a baseline. Second, we initialize from pretrained unimodal models (RoBERTa and ViT, \citealp{Liu2019RoBERTaAR,DosovitskiyB0WZ21}), which can potentially reduce the amount of bimodal pretraining. Last, we pretrain the entire VL encoder with bimodal supervision (Figure~\ref{fig:overview}, Center), and compare different data sources for pretraining, each requiring different amounts of human effort.

In each experiment we finetune and evaluate the VL encoder on downstream VL applications (Figure~\ref{fig:overview}, Right), focusing on classification tasks (visual question answering and visual reasoning). 

\section{Data}

We now describe the datasets used during bimodal pretraining and finetuning. For downstream applications, we put an emphasis on tasks that require reasoning over image(s) and text.  Table~\ref{tab:datasets_examples} provides an example from each dataset, and Appendix~\ref{sec:training_data} provides key statistics and details on the composition of the training sets.
%\footnote{For some datasets the number of training instances in Table~\ref{tab:datasets_stats} is slightly lower than stated in the original work due to either broken image URLs or filtering we performed (described in the paper text).}

\begin{table*}[h]
\centering
\tiny
%\hspace*{-1.7cm}
\begin{tabular}{ccc}
\toprule
\quad\quad\quad\quad\quad\quad ImageNet \quad\quad\quad\quad\quad\quad & Conceptual Captions & Conceptual Captions Image Labels \\ 
\midrule
\raisebox{-25pt}{\makecell{\includegraphics[scale=0.25]{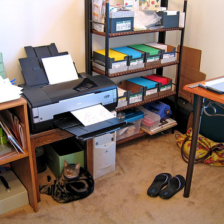}}} & \ 
\raisebox{-25pt}{\makecell{\includegraphics[scale=0.25]{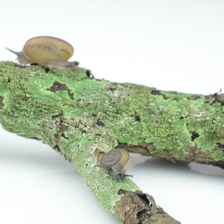}}} & \ 
\raisebox{-25pt}{\makecell{\includegraphics[scale=0.25]{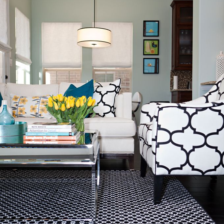}}} \\
\multicolumn{1}{l}{\makecell[l]{\textit{Class label}: printer}} & 
\multicolumn{1}{l}{\makecell[l]{\textit{Caption}: snail on a branch isolated \\ on white background}} & 
\multicolumn{1}{l}{\makecell[l]{\textit{Computer-generated labels}: room, interior design,\\ furniture, blue, living room, green, property, turquoise,\\ home, floor, yellow, table, building, wall, house
}} \\
%\bottomrule
\end{tabular}
\vskip 2pt
%\hspace*{-1.7cm}
\begin{tabular}{ccc}
\toprule
VQAv2 & GQA & NLVR2 \\ 
\midrule
\raisebox{-25pt}{\makecell{\includegraphics[scale=0.25]{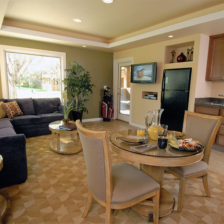}}} & \ 
\raisebox{-25pt}{\makecell{\includegraphics[scale=0.25]{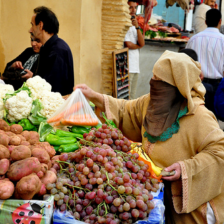}}} & \ 
\raisebox{-25pt}{\makecell{\includegraphics[scale=0.25]{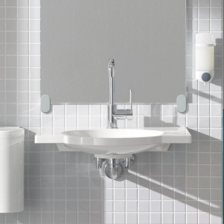} \includegraphics[scale=0.25]{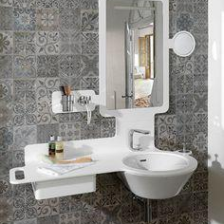}}} \\
\multicolumn{1}{l}{\makecell[l]{\textit{Question}: How many chairs can \\ you count? \textit{Answer:} 2}} & 
\multicolumn{1}{l}{\makecell[l]{\textit{Question}: What vegetable is to the left \\ of the bag? \textit{Answer:} cauliflower}} & 
\multicolumn{1}{l}{\makecell[l]{\textit{Sentence}: The sink in one of the images is set into \\ a brown wood hanging counter. \textit{Label:} false}} \\
\bottomrule
\end{tabular}
\vspace{-0.3cm}
\caption{Examples from all datasets used in this work.}
\label{tab:datasets_examples}
\end{table*}

\subsection{Pretraining Datasets}
\paragraph{ImageNet-1K \cite{ILSVRC15}}
is a human-annotated dataset that consists of over 1.2M images, divided into 1,000 classes that are mapped to meaningful concepts according to the WordNet hierarchy \cite{wordnet}. Each concept is described by one or more language phrases, and accompanied by $\sim$1000 images to illustrate it.
We consider ImageNet-1K as a source of lightweight bimodal supervision, relatively cheap to obtain, as images are paired with text describing a single concept rather than a full-sentence.

\paragraph{Conceptual Captions (CC) \cite{sharma2018conceptual}}
is a programmatically-generated dataset of image-text pairs that consists of 3.3M examples.
Prior work has demonstrated that CC is an effective resource for vision-language pretraining \cite{pmlr-v139-kim21k,ALBEF,lu2019vilbert,hendricks-etal-2021-decoupling}.
We use CC as a primary source of bimodal supervision, since: (a) it does not involve manual annotations, (b) it is small enough to be used by resource-constrained researchers, and (c) its images are from a different origin than the downstream tasks. Therefore, it provides a suitable test bed for estimating models' ability to generalize to new images.

\paragraph{Conceptual Captions Image Labels (CCIL) \cite{ng-etal-2021-understanding}}
is a subset of 2M images from CC that contains machine-generated labels using the Google Cloud image labelling API. 
While labels are cheap since they are automatically-generated, the API was presumably trained on large amounts of bimodal data. Nevertheless, we examine pretraining on images paired with sets of labels to investigate whether this provides a sufficiently rich source of bimodal supervision despite lacking natural language sentences. 
Past work indeed showed that VL pretraining benefits from masking object labels \cite{bitton-etal-2021-data-efficient}.

\subsection{Downstream Tasks}\label{sec:downstream_tasks}
\paragraph{VQAv2 \cite{Goyal2018MakingTV}} 
VQAv2 is a human-authored visual question answering (VQA) dataset that consists of 1.1M natural language questions with 10 short answers per question over 204K images from COCO \cite{Lin2014MicrosoftCC}.
%To counter language bias, each question appears twice but paired with two similar images that lead to different answers.
It is standard to treat VQAv2 as a classification task, by only keeping questions with the most common answers (3,129 classes) \cite{Anderson2017up-down, tan-bansal-2019-lxmert, Zhai_2022_CVPR}.

\paragraph{GQA \cite{Hudson2019GQAAN} (balanced)} 
is a VQA dataset whose public version contains 1.1M questions over 83K images from Visual Genome~\cite{krishnavisualgenome}. Unlike VQAv2, questions are created programmatically from scene graphs created by human annotators.
Using scene graphs allows GQA to generate questions that test various reasoning skills such as comparisons, logical inference, spatial reasoning, etc.

\paragraph{NLVR2 \cite{suhr-etal-2019-corpus}}  
is a benchmark for testing models' ability to reason over text and images. The dataset contains 107K examples, where each example contains an English sentence and two web images (see Table~\ref{tab:datasets_examples}). The goal is to determine whether the sentence is true or false in the context of the pair of images, a binary classification task.

\section{Method}

Our goal is to develop a classifier $f: \mathcal{X} \times \mathcal{I} \rightarrow \mathcal{C}$ that given an utterance $\bmx$ and an image $\bmi$ predicts a class $c \in \sC$.

\subsection{Architecture}
We use a VL architecture, adapted from \citet{Dou_2022_CVPR}.
The tokens of the utterance $\bmx = (x_0, x_1, \dots, x_n)$ are fed into a transformer \emph{text encoder}, where $x_0$ is the special symbol $\texttt{CLS}_\text{txt}$. Similarly, the image is broken into patches $\bmi = (i_0, i_1, \dots, i_m)$, where $i_0$ is a special symbol $\text{CLS}_\text{img}$, which are fed into a transformer \emph{image encoder}.

The image and text encoders compute contextualized representations of the image and text $(\hat{h}_0^{\text{txt}}, \dots,  \hat{h}_n^{\text{txt}})$ and $(\hat{h}_0^{\text{img}}, \dots,  \hat{h}_m^{\text{img}})$, which are then
linearly projected with projection matrices $W_\text{proj}^\text{txt} \in \mathbb{R}^{d_\text{txt} \times d}, W_\text{proj}^\text{img} \in \mathbb{R}^{d_\text{img} \times d}$, where $d_\text{txt}, d_\text{img}$ are the hidden state dimensions of the text and image encoders respectively. 
The projected representations of each modality are then passed through transformer \emph{fusion} layers, which include both a self-attention sublayer, and a cross-attention sublayer. Namely, each modality performs cross-attention on the other modality to fuse information from its representations, capturing interaction between the modalities.
Overall, the VL encoder outputs the image-and-text contextualized representations 
$\bm{h}^\text{img} = (h_0^{\text{img}}, \dots, h_n^{\text{img}})$ and
$\bm{h}^\text{txt} = (h_0^{\text{txt}}, \dots, h_m^{\text{txt}})$.
An overview of our architecture is given in Figure~\ref{fig:overview}, Left.

All model parameters are jointly trained by defining loss functions over classification heads, which we describe next. 
Since some model parameters are initialized from a pretrained model, while other are randomly initialized, we use a higher learning rate for randomly initialized weights compared to pretrained weights, similar to \citet{Dou_2022_CVPR}.

\subsection{Pretraining Objectives}\label{subsec:pretrain_objectives}

For pretraining, we use two objectives: masked language modeling (MLM) \cite{devlin-etal-2019-bert, tan-bansal-2019-lxmert}, and image-text matching (ITM) \cite{tan-bansal-2019-lxmert}, which are the most common objectives for VL pretraining and lead to state-of-the-art performance \cite{Dou_2022_CVPR}.
During training, we sum the ITM loss and the MLM loss for each training instance.

In MLM, given a masked token $x_i$ the goal is to maximize the probability of the gold token given the representation $h_i^\text{txt}$, using cross-entropy loss. In ITM, given a image-text pair $(\bmx, \bmi)$, we concatenate the special CLS tokens $h_0^\text{img}$ and $h_0^\text{txt}$, and
use a sigmoid layer to predict if the image matches the text or not. We train with binary cross-entropy loss.

When pretraining on Conceptual Captions, we use the same masking scheme employed by \citet{Dou_2022_CVPR}, that is, randomly masking 15\% of the tokens. 
For ImageNet, we are given an image and a text label and mask all of its tokens. For CCIL, we are given an image and a list of text labels, concatenated with commas as separators, ordered by their machine-generated confidence scores. We then mask all tokens of a randomly-sampled label.

In ITM, in 50\% of the examples, given a positive pair $(\bmx, \bmi)$, we substitute the true image with a random one and label it as a negative example.

\subsection{Finetuning}

Since the downstream applications in \S\ref{sec:downstream_tasks} can all be framed as classification tasks, we add a classification head to finetune the VL encoder. The classification head is a two-layer MLP, as in \citet{pmlr-v139-kim21k}. Specifically, we take as input the concatenation of all the image and text \texttt{CLS} representations, i.e.,  $[h^{\text{img}}_0; h^{\text{txt}}_0]$, and use the MLP to map them to $|\sC|$ logits based on the number of task classes.
The objective during training is to maximize the probability of the correct class(es), and we use standard cross-entropy loss.
At inference time, we return the top-scoring class for all downstream tasks.

In NLVR2, where each example has two images, we consider each example as two image-text pairs, duplicating the text, and pass them separately through the VL encoder (dubbed `the pair setup' in \citet{chen2020uniter}). We then pass four \texttt{CLS} representations (two for the images, two for the text) to the MLP to obtain the prediction.

\section{Experiments}

%We now present experiments which test performance on vision-language downstream tasks given varying amount of bimodal supervision.

\paragraph{Experimental Setup}
We use ViT-Base \citep{DosovitskiyB0WZ21} as the image encoder, pretrained and finetuned on ImageNet-21K at a resolution of 224x224 with a patch size of 16x16. We use RoBERTa-Base \citep{Liu2019RoBERTaAR} as the text encoder. For the cross-modal transformer, we use only two layers to save computational resources, as previous work \citep{lu2019vilbert,hendricks-etal-2021-decoupling}, as well as our own preliminary findings, have shown that the effect of depth is small after finetuning. 

We run pretraining (\S\ref{subsec:pretrain_objectives}) for a maximum of 7,400 steps, and finetune each downstream task for 10 epochs. We specify batch sizes and learning rates for each case in Appendix \ref{subsec:implementation_details}.

The evaluation score for VQAv2 is VQA score, and accuracy for GQA and NLVR2.
Each result for VQAv2 and GQA is a 3-run average on the test-dev split, and for NLVR2 a 10-run average on the public test split.

\paragraph{Limitations}
%\section{Limitations}
Our work is performed within a limited compute budget. Therefore, we choose our largest pretraining dataset to be CC although there are datasets orders of magnitude larger. 
Compared to other work, we use images in a lower resolution, which has been shown to decrease performance \cite{Dou_2022_CVPR}. Also, \citet{Dou_2022_CVPR} showed that better image encoders can significantly improve performance even before bimodal pretraining, but we did not experiment with different text and image encoders nor with larger models. Additionally, even though further pretraining in some setups results in small performance improvements, we decided the computational cost was unjustified. \citet{bugliarello-etal-2021-multimodal} showed pretraining variance exists when training on CC, but we were only able to pretrain once in each setup, due to the high computational costs. All of the above means that our work is self-contained, but cannot be directly compared in numbers to other works.

\subsection{Main Results}
\label{subsec:main_results}

\begin{table*}[h]
\centering
\footnotesize
\begin{tabular}{llll}
\toprule
                      &      VQAv2 &        GQA &          NLVR2 \\
\midrule
         Random init. &          52.3$\pm$0.1 &          43.1$\pm$0.1 &          random \\
  Vision Random init. &          54.2$\pm$0.0 &          44.3$\pm$0.2 &          random \\
Language Random init. &          66.3$\pm$0.1 &          51.2$\pm$0.1 &          random \\
Unimodally-pretrained &          69.5$\pm$0.1 &          53.6$\pm$0.1 &          random \\
Bimodally-pretrained with CCIL & 69.6$\pm$0.3 &          52.9$\pm$0.5 &          68.0$\pm$0.7 \\
Bimodally-pretrained with CC & \textbf{70.7$\pm$0.0} & \textbf{56.1$\pm$0.3} & \textbf{74.3$\pm$0.3} \\
\bottomrule
\end{tabular}
\vspace{-0.3cm}
\caption{Main results for all downstream tasks.}
\label{table:main_results}
\end{table*}

Table \ref{table:main_results} shows the results of finetuning on all downstream tasks for different initializations of the image and text encoders.

In addition to finetuning a model that is initialized with ViT and RoBERTa (`Unimodally-pretrained'), and in order to verify the importance of unimodal pretraining, we finetune our model when the image encoder, text encoder, or both encoders are randomly initialized. We find that pretraining the vision model is essential for good performance, and observe a smaller drop in performance when the text encoder is randomly initialized, similar to \citet{Zhai_2022_CVPR}.

Comparing the unimodally-pretrained model to one that was further pretrained on CC (`Bimodally-pretrained with CC'), we see that for VQAv2 the gap is only 1 point, and for GQA it is just 2.5 points. However, on the more challenging NLVR2, which requires complex reasoning operations, bimodal pretraining is essential, and the model achieves random performance without it. Nevertheless, training with a weaker form of supervision, namely, a list of machine-generated object labels from CCIL is sufficient for non-random and reasonably high performance on NLVR2 (but has no effect on VQAv2 and GQA).

\subsection{Effect of CC Size on Pretraining}
\label{subsec:cc_pretraining}
Table~\ref{table:main_results} showed that bimodal pretraining is essential for obtaining non-random results on NLVR2. A natural question is whether this can be obtained with fewer pretraining examples. To this end, we pretrain on different fractions of CC and present the results after finetuning in Table~\ref{table:pretraining_results} (in the Appendix) and Figure~\ref{fig:cc-size}.

Surprisingly, even when using only $\sim$1\% of CC (30K examples), performance on NLVR2 is far from random -- 67.3. When using 5\% of the data, performance is only moderately lower than when using CC in its entirety -- 70.2 vs. 74.3. When using 25\% of the data for pretraining, performance on all three datasets is less than two points lower than when using 100\%, showing that indeed the amount of bimdal supervision can be considerably reduced with only a small hit in performance.

\begin{figure}
\begin{floatrow}
\ffigbox[]{%
  \includegraphics[scale=0.335]{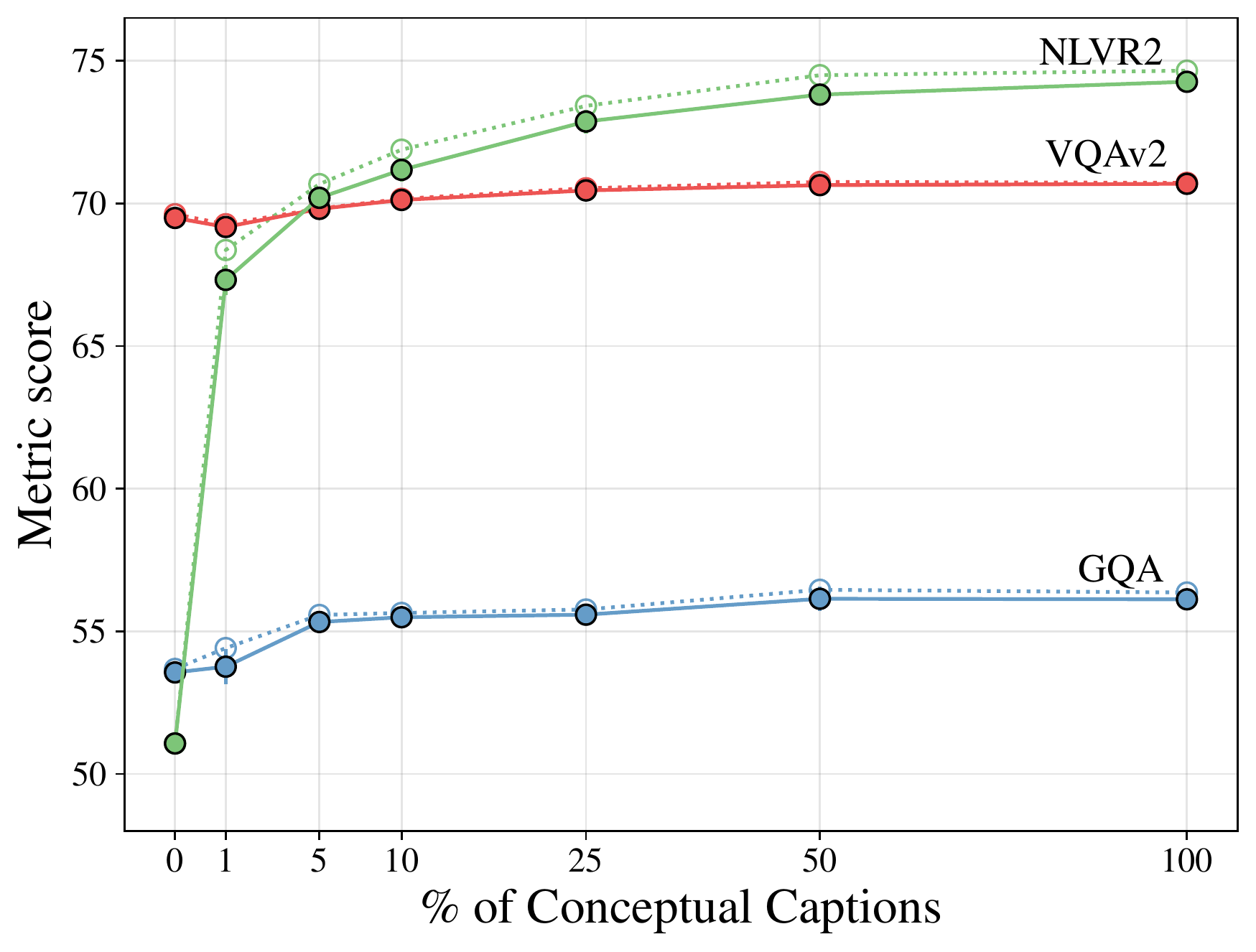}%
}{%
\vspace{-0.4cm}
  \caption{Effect of the fraction of examples from CC on downstream task performance. Solid/dashed line -- average/maximum score over seeds.} \label{fig:cc-size}%
}%
\capbtabbox[]{%
\centering
\footnotesize
  \begin{tabular}{l|ll}
\toprule
Max \# of labels & Unique Labels &                 NLVR2 \\
\midrule
              1 &  5.3K          &          52.6$\pm$1.4 (55.3) \\
              3 &  8.0K          &          67.8$\pm$0.5 (68.6) \\
             15 &  14.3K         & \textbf{68.0$\pm$0.7 (68.9)} \\
\bottomrule
\end{tabular}
\begin{tabular}{lll}
              &                  &                             \\
\end{tabular}%
}{%
\vspace{-0.4cm}
  \caption{Performance on NLVR2 when restricting the number of labels per image during pretraining on CCIL (max. value is in the parentheses).} \label{table:ccil_results}%
}
\end{floatrow}
\end{figure}

The aforementioned results were obtained by finetuning on all of the downstream data per task. However, an interesting variant to consider is a low-resource setting where we have only \emph{some} of the downstream data -- what is the importance of bimodal pretraining then? Table~\ref{table:low_resource_downstream} (in the Appendix) and Figure~\ref{fig:low_resource_downstream} show for VQAv2 and GQA that when less data is used for finetuning, the benefit of pretraining with 5\% or more of CC is greater than the benefit observed when 100\% of the downstream data is used for finetuning. For NLVR2, we see that pretraining is still very helpful even with 100\% of the downstream data. The reason for the difference might be that VQAv2 and GQA are much larger than NLVR2.

\begin{figure}[h]
   \centering
   \includegraphics[scale=0.2877]{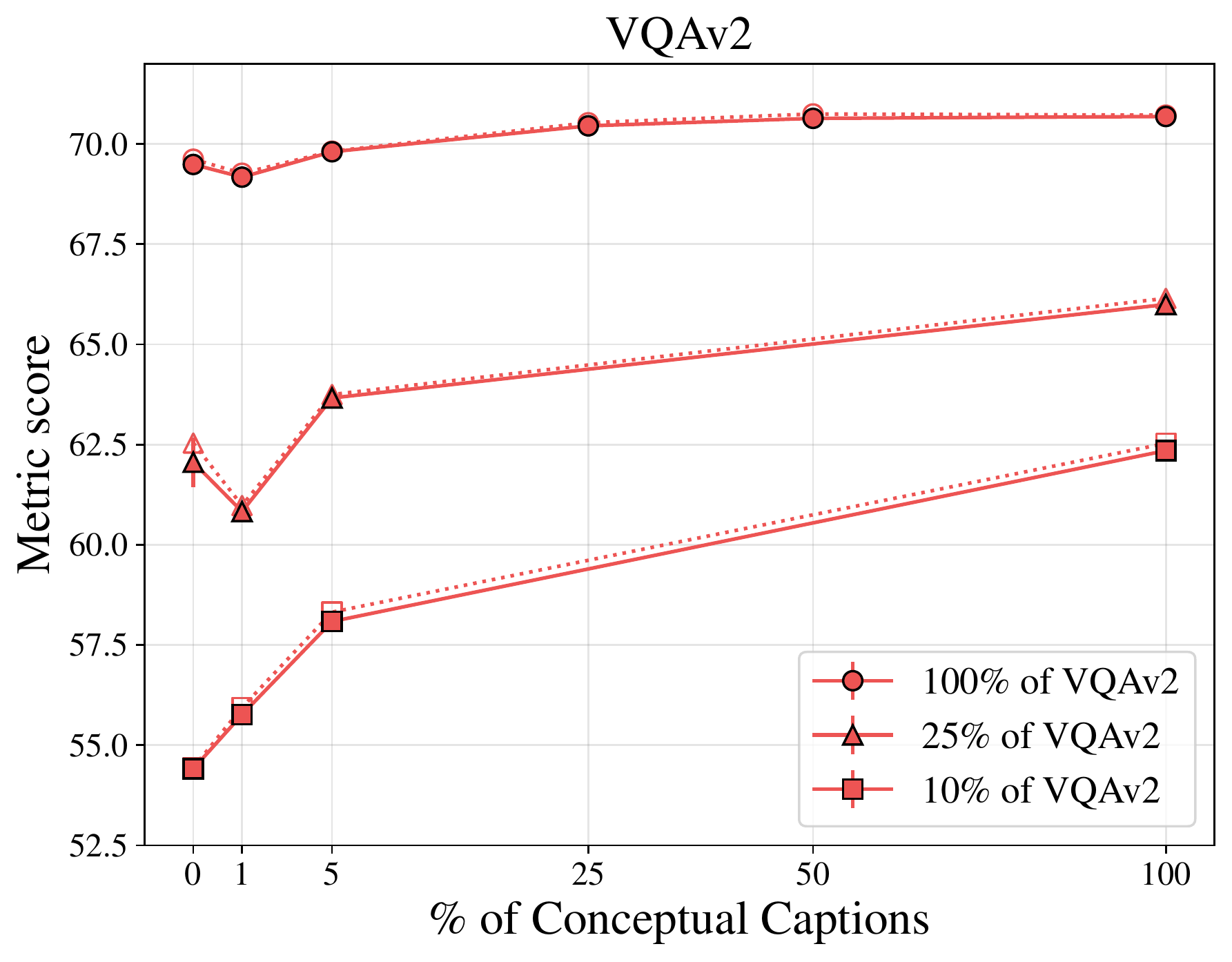}
   \includegraphics[scale=0.2877]{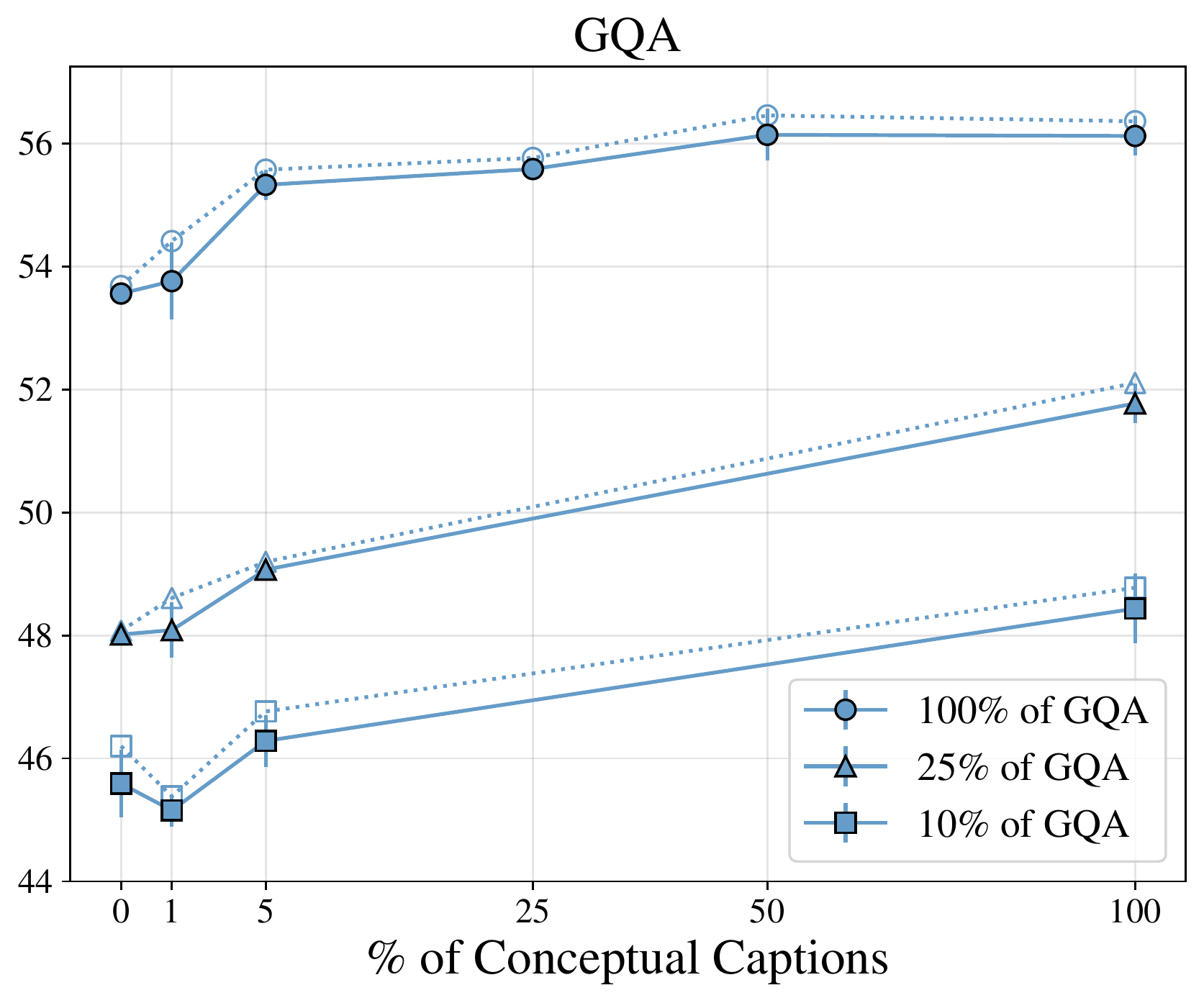}
   \includegraphics[scale=0.2877]{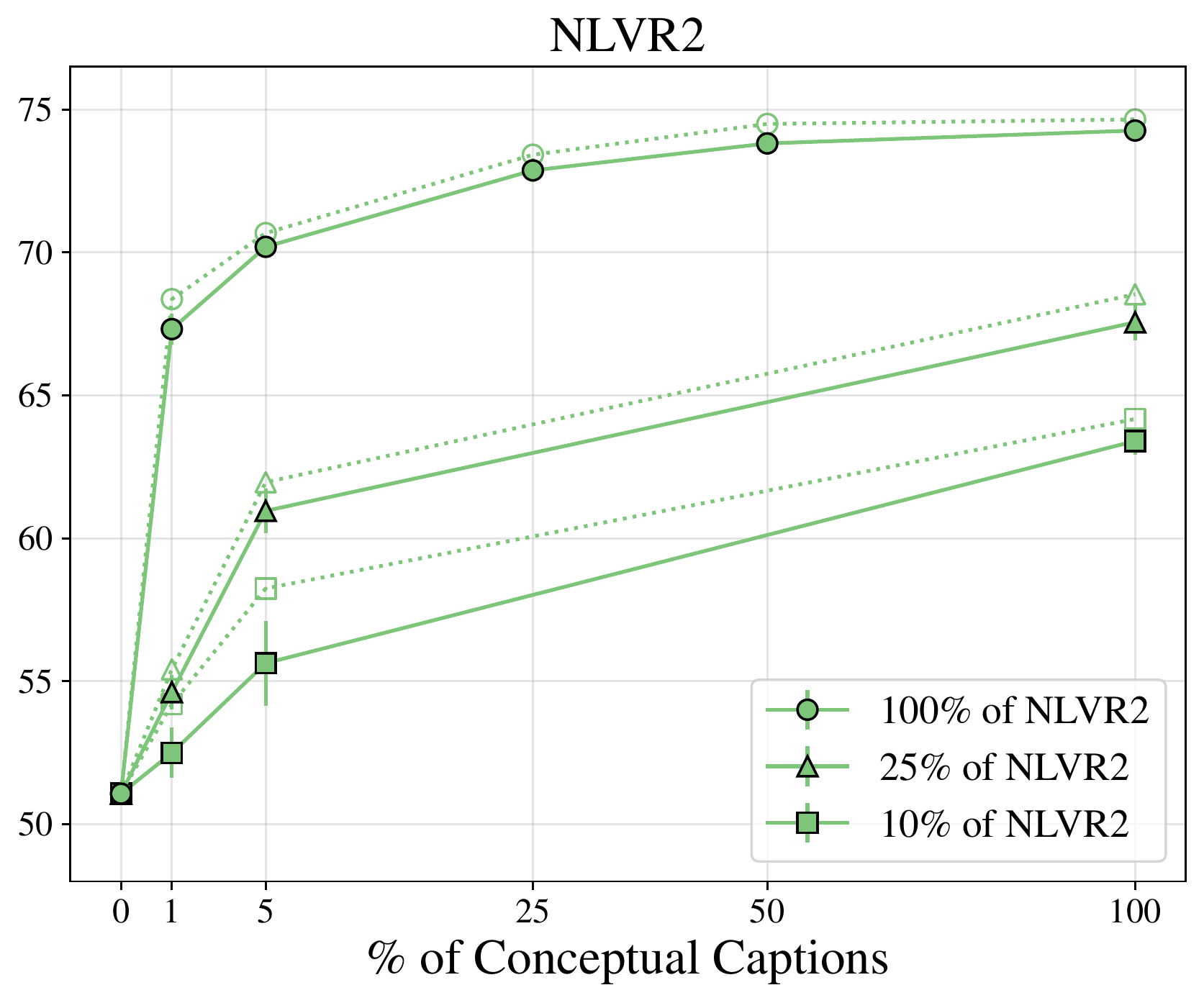}
   \vspace{-0.75cm}
  \caption{Effect of the fraction of examples from CC on downstream task performance when finetuning on less downstream data. Solid/dashed line -- average/max. over seeds.} \label{fig:low_resource_downstream}
\end{figure}

\subsection{Pretraining with ImageNet Labels}
We have seen in \S\ref{subsec:main_results} that image-caption pairs are useful for pretraining VL models. Here, we investigate if a weaker source of language supervision, namely image labels only, suffices for aligning text and vision representations.
Specifically, we pair each ImageNet image with its label, treating it as a caption, and pretrain with MLM and ITM as described in \S\ref{subsec:pretrain_objectives}.

We observe \emph{no difference} in results compared to unimodally-pretrained models (Table~\ref{table:imagenet_results} in the Appendix) -- performance remains random for NLVR2, and similar for VQAv2 and GQA. This suggests that ImageNet labels do not provide adequate signal for VL pretraining.

\subsection{Pretraining with CCIL}
\label{subsec:ccil_pretraining}
One hypothesis for the lack of improvement when pretraining with ImageNet is that a single label per image is too limiting, since images typically contain many objects. To test this, we pretrain with CCIL, where each image is paired with machine-generated labels, providing a richer image representation. We pretrain with MLM and ITM as described in \S\ref{subsec:pretrain_objectives}.

While pretraining on CCIL does not improve performance on VQAv2 and GQA, it leads to dramatic improvement on NLVR2, reaching an average accuracy of  68.0$\pm$0.7 and a maximum accuracy of 68.9. This shows that providing a set of object labels lets the model better align image and text representations. 
Table \ref{table:ccil_results} further validates this by showing results when restricting the maximal number of labels per image. We observe that having multiple labels per image is crucial, as performance is roughly random when using a single label. Using 3 labels is already sufficient for bootstrapping the model, and performance is barely lower compared to using all 15 labels.

\subsection{Transfer Learning}
Finally, we test whether a model finetuned on a source downstream task (VQAv2 and GQA) can improve performance on a target task, i.e., in a transfer learning setup, where we vary the amount of annotated data in the source task.

Table \ref{table:vqa_pretraining_results} (in the Appendix) and Figure~\ref{fig:transfer} (left) show results when VQAv2 is the source task and GQA and NLVR2 are the target tasks. VQAv2 appears to be an effective source of bimodal supervision for both tasks -- when using all of VQAv2, performance on GQA is even slightly higher compared to pretraining on CC data, and 3 points lower on NLVR2 (74.3$\rightarrow$71.1). Nevertheless, the amount of data in the source task is important, and performance on NLVR2 is much lower when using 5\%-25\% of the data.

Table \ref{table:gqa_pretraining_results} (in the Appendix) and Figure~\ref{fig:transfer} (right) show results when GQA is the source task and VQAv2 and NLVR2 are the target tasks. We observe that VQAv2 is a better source task compared to GQA -- GQA does not improve performance on VQAv2, and its effect on NLVR2 is much more moderate. A possible explanation is that VQAv2 has natural language questions, while questions in GQA are automatically generated. Another potential factor is the fact that VQAv2 typically require less reasoning steps compared to GQA.

Overall, in both cases we find transfer learning on downstream tasks is useful, and can even perform closely to bimodally-pretrained models.

\begin{figure}[t]
   \centering
   \includegraphics[scale=0.332]{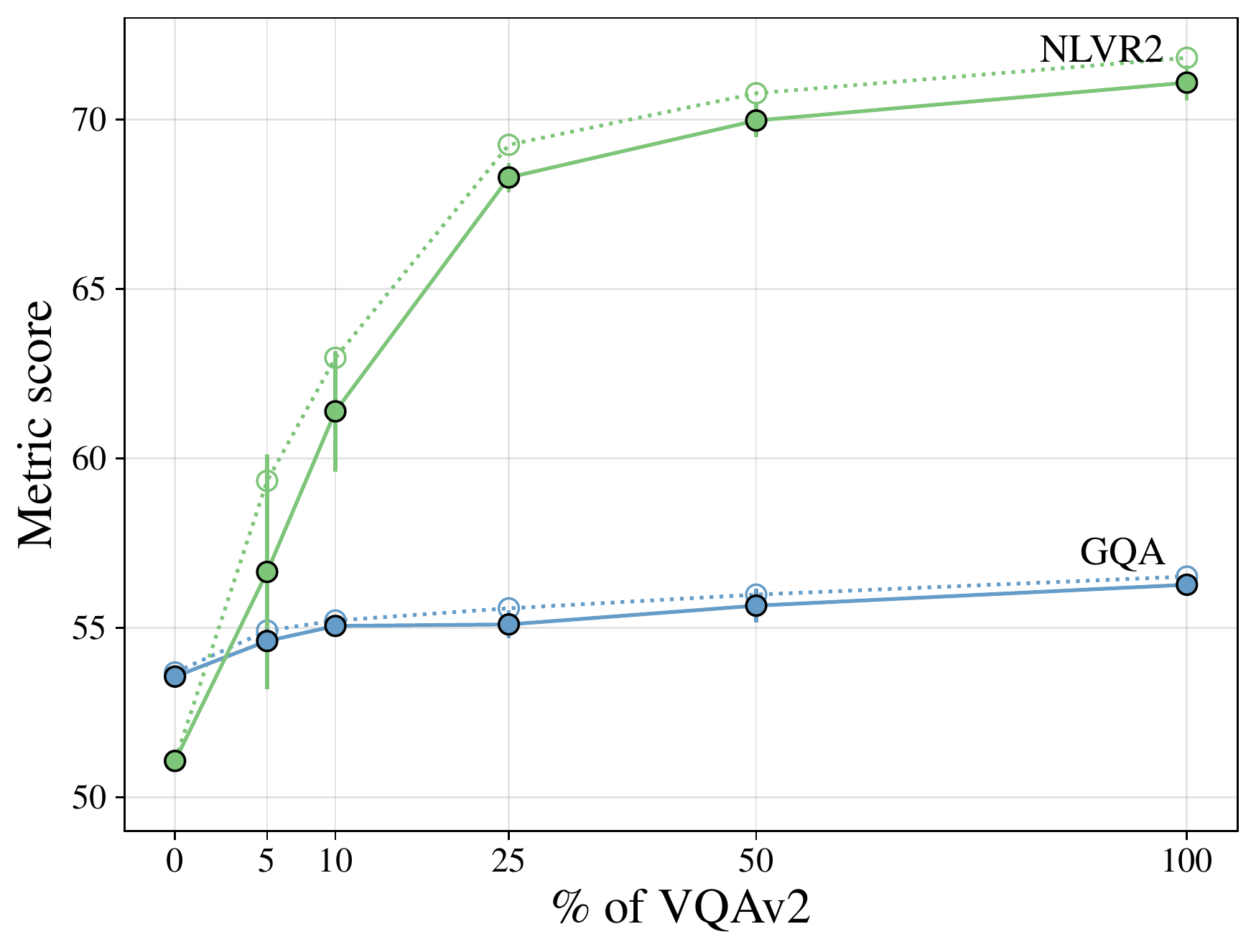}
   \quad
   \includegraphics[scale=0.332]{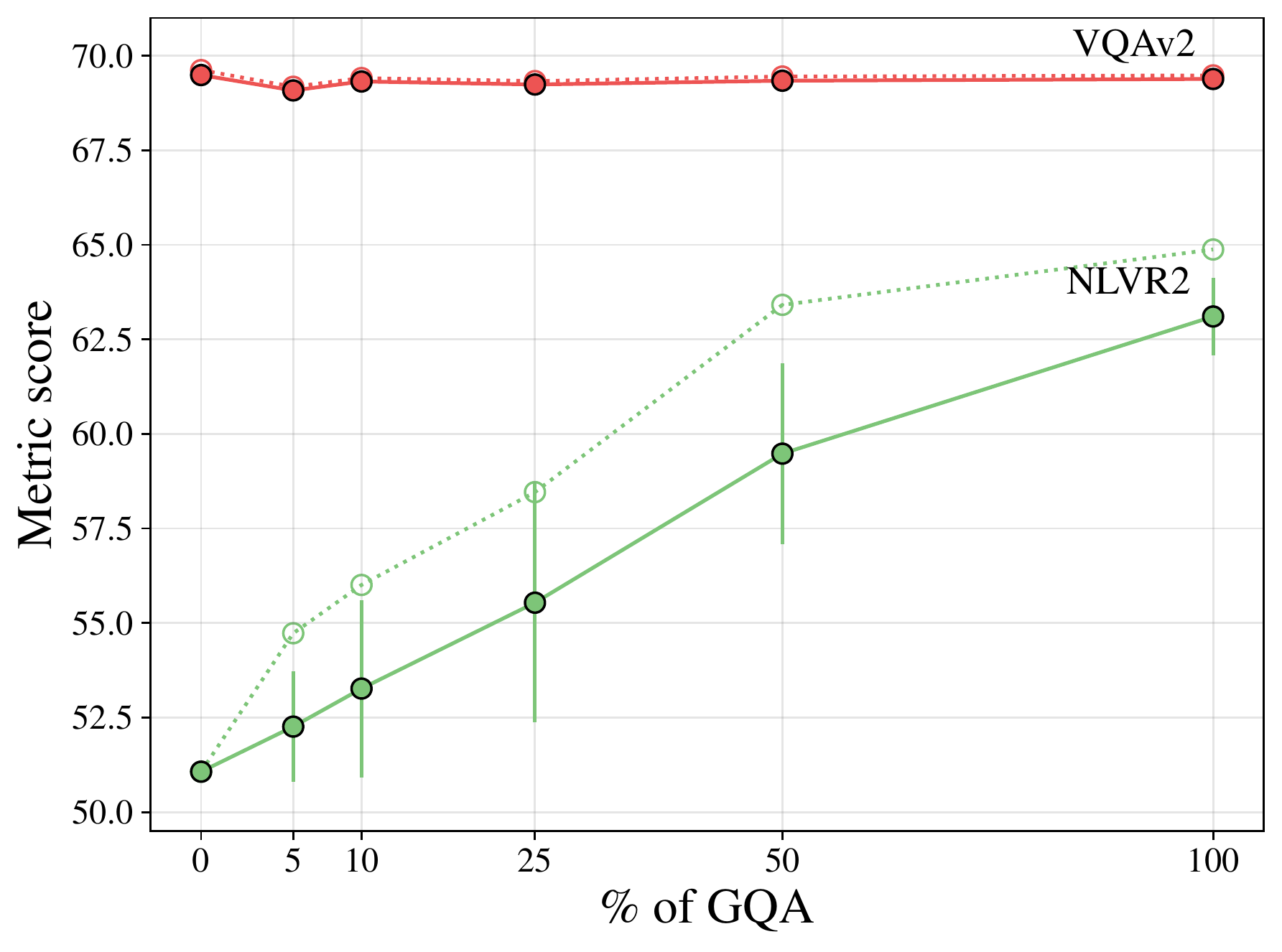}
   \vspace{-0.4cm}
  \caption{Effect of the fraction of examples from VQAv2 (left) and GQA (right) on downstream task performance. Solid/dashed line -- average/maximum score over seeds.} \label{fig:transfer}
\end{figure}

\section{Analysis}
To better understand what data properties are important for pretraining, we train on small subsets of CC (1\% of the data) and VQAv2 (5\% of the data), with particular characteristics:
\begin{itemize}[leftmargin=*,topsep=3pt,itemsep=3pt,parsep=0pt]
    \item \textit{Min/max length}: We create subsets that minimize/maximize the average input length.
    \item \textit{Min/max vocabulary size} - 
    We create subsets that minimize/maximize the size of the vocabulary. To do so we use a greedy procedure, where (a) we initialize an empty set of examples, and at each step (b) randomly sample a candidate set of 10K examples, and (c) choose the example that minimizes/maximizes the current vocabulary size.
\end{itemize}

\begin{table}[ht]
\centering
\footnotesize
\begin{tabular}{l|lll|lll}
\toprule
\multicolumn{1}{c|}{} & \multicolumn{3}{c|}{1\% CC} & \multicolumn{3}{c}{5\% VQAv2} \\
\midrule
Subset & Length & Vocab. &                        NLVR2 & Length & Vocab. &                        NLVR2 \\
\midrule
   Min length &  5.0 &  8.0K &          67.1$\pm$0.3 (67.4) &  4.45 &  3.6K &          57.4$\pm$3.1 (60.8) \\
   Max length &  30.3 &  19.1K &          64.1$\pm$1.4 (65.7) &  12.7 &  7.6K &          53.9$\pm$2.0 (57.0) \\
   Min vocab. &  6.5 &  0.3K &          64.8$\pm$1.2 (65.9) &  5.8 &  0.2K & \textbf{57.4$\pm$3.8 (62.2)} \\
   Max vocab. &  14.0 &  44.4K & \textbf{67.3$\pm$0.3 (67.7)} &  7.7 &  16.4K &          55.1$\pm$3.1 (58.2) \\
       Random &  10.3 &  13.3K &          67.3$\pm$0.5 (68.4) &  7.3 &  5.8K &          56.6$\pm$3.5 (59.3) \\
\bottomrule
\end{tabular}
\vspace{-0.4cm}
\caption{Analyzing the effect of pretraining on CC/VQAv2 subsets with particular properties. After training on each subset, we finetune on NLVR2.}
\label{table:analysis_results}
\end{table}

\vspace{-0.3cm}
Results are in Table ~\ref{table:analysis_results}.
No subset is noticeably better than a random subset. For CC, results are similar. For VQAv2, while performance when minimizing length and vocabulary is better on average, the differences seem negligible, given the high standard deviation.

\paragraph{Effect of length on pretraining}
Table~\ref{table:analysis_results} shows that pretraining on long inputs substantially hurts performance -- results are reduced by at least 3 points for both CC and VQAv2.
This is surprising as one might hypothesize that longer inputs should be better since they contain more information.
A possible explanation is that simple examples are necessary to bootstrap the pretraining procedure and align the text and image representations.

\paragraph{Effect of vocabulary size on pretraining}
Pretraining on a subset with higher lexical diversity should expose the model to more concepts, both in images and texts, and therefore improve its performance. While for CC this is indeed the case, for VQAv2 results for the max vocabulary size setup with 16.4K words are lower than the min vocabulary size setup with only 0.2K words. A possible explanation is the amount of yes/no questions in the min/max vocabulary size subsets which is 80.7\% and 44.5\%, respectively --  Since NLVR2 is a yes/no task, training on more yes/no questions might be closer to its distribution. 

\vspace{-0.024cm} 
\section{Related Work}
\citet{Dou_2022_CVPR} investigated unimodally-pretrained models, finetuning different image and text encoders on multiple VL tasks, recognizing it as efficient and performant. 
However, they did not consider the effects of the types and amount of bimodal supervision.
Past work investigated bimodal supervision on VL models, but for models that use \emph{frozen} features from an object detection model \cite{Singh2020AreWP,hendricks-etal-2021-decoupling}, which (a) cannot be adapted to unseen concepts \cite{Zhang_2021_CVPR}, (b) require heavily annotated object-level data for the training of the object detection model \cite{krishnavisualgenome,Anderson2017up-down}, and (c) result in an architectural inductive bias towards objects (which is very beneficial for VQA tasks).
\citet{Singh2020AreWP} compared performance between multiple pretraining datasets, varying their sizes. Unlike us, for all tasks, the effect of different usage of bimodal supervision was small, compared to our NLVR2 experiments. \citet{hendricks-etal-2021-decoupling} assessed the contribution of pretraining datasets from a set of standard VL pretraining datasets, but focused on zero-shot image retrieval tasks. 

\citet{li-etal-2021-unsupervised} and \citet{Zhou_2022_CVPR} also share the motivation of reducing bimodal pretraining for VL models. With some similarity to our CCIL experiments in \S\ref{subsec:ccil_pretraining}, they avoid pretraining on collected parallel image-text data altogether by utilizing predictions of regions and tags from an object detection model to create VL-specialized training objectives.

Opposite to our setup, a current trend is to pretrain models on vast amounts of bimodal data \cite{radford2021learning, Zhai_2022_CVPR, Alayrac2022FlamingoAV}, and perform zero/few-shot evaluation. While remarkable results were achieved, performance is lower than finetuned models pretrained on less bimodal data, which is relatively cheap to obtain.

\vspace{-0.024cm} 
\section{Conclusion}
A current obstacle on the road to multimodal models is reliance on bimodal supervision.
In this work, we go in an opposite direction from current trends, and instead of using increasing amounts of bimodal data, we examine whether one can use \textit{less} of it.
We find that indeed this is the case, where for simple tasks just finetuning unimodally-pretrained models leads to performance that is similar to bimodally-pretrained models, at a much lower cost.
For complex tasks, while bimodal pretraining is still necessary, its amount (100\%$\rightarrow$5\%) and source quality (CC$\rightarrow$CCIL) can be significantly reduced with only a moderate degradation in performance. We also find that models finetuned on one downstream task are useful in a transfer learning setup, achieving results close to bimodally-pretrained models.

\vspace{-0.024cm} 
\section*{Acknowledgements}
This research was partially supported by The Yandex Initiative for Machine Learning, and the European Research Council (ERC) under the European Union Horizons 2020 research and innovation programme (grant ERC DELPHI 802800).

\bibliography{sample2e}
\bibliographystyle{plainnat}

\newpage
\appendix
\section*{Appendix for ``Training Vision-Language Models with Less Bimodal Supervision''}

\section{Training Data}\label{sec:training_data}
Since for some of datasets the official training splits aren't used as-is, we provide the exact details of the training data composition for each dataset and also key statistics for all of the datasets in Table~\ref{tab:datasets_stats}.

\begin{table*}[h]
\centering
\footnotesize
\begin{tabular}{lccc}
\toprule
Dataset  & Training instances    & Unique texts      & Training images  \\ 
\midrule
ImageNet & 656K & 738 & 656K \\ 
Conceptual Captions (CC) & 2.84M & 2M & 2.84M \\ 
Conceptual Captions Image Labels (CCIL) & 1.84M & 1.79M & 1.84M \\ 
VQAv2 & 620K & 210K & 118K \\ 
GQA    & 943K  & 538K & 72K \\ 
NLVR2    & 86K  & 23K & 103K \\ 
\bottomrule
\end{tabular}
\vspace{-0.3cm}
\caption{Key statistics for the training datasets.}
\label{tab:datasets_stats}
\end{table*}

\paragraph{ImageNet-1K \cite{ILSVRC15}} Since ImageNet classes are often too fine-grained, we manually collapse fine-grained classes into an ancestor WordNet class,\footnote{\url{https://observablehq.com/@mbostock/imagenet-hierarchy}.} e.g., dog breeds are collapsed to ``dog''. Then, we create a balanced training set according to the updated classes of the images.

Following is a list of the classes we collapse sub-classes into:\\
\texttt{dog, fox, wild dog, wolf, coyote, domestic cat, bear, monkey, snake, lizard, turtle, frog, salamander, lobster, crab, beetle, butterfly, spider, rabbit, bird, fungus}

\paragraph{Conceptual Captions (CC) \cite{sharma2018conceptual}}
Out of the 3.3M examples in the official training set, we were able to download 2.84M examples from the provided image URLs.

\paragraph{Conceptual Captions Image Labels (CCIL) \cite{ng-etal-2021-understanding}}
Out of the 2M examples in the official training set, we were able to download 1.84M examples from the provided image URLs.

\paragraph{VQAv2 \cite{Goyal2018MakingTV}} 
We create our training set similar to previous works on VQAv2 \cite{tan-bansal-2019-lxmert,Dou_2022_CVPR}, and use the same validation set as \citet{tan-bansal-2019-lxmert}, which was constructed from the official validation set based on 5,000 randomly chosen images.

To create the training set, we first create an answer set that contains only majority answers that occurred at least 9 times on the official training and validation sets together.
Then, out of the official training and validation sets, we filter out all of the examples that doesn't have any answer in the created answer set.
Finally, out of the remaining examples, we discard every example that appears in our validation set.

\paragraph{GQA \cite{Hudson2019GQAAN}}
We use the official training set.

\paragraph{NLVR2 \cite{suhr-etal-2019-corpus}} 
We use the official training set.

\section{Experimental Setup}\label{sec:appendix_experimental}

\subsection{Additional Implementation Details}\label{subsec:implementation_details}
\paragraph{Image Preprocessing} Both in pretraining and finetuning, we apply center crop on the image and resize it to 224x224. When training, we additionally use RandAugment \cite{NEURIPS2020_d85b63ef} as in \citet{pmlr-v139-kim21k} with the exclusion of color-changing strategies (Invert, Posterize, Solarize, SolarizeAdd) and the coutout strategy.

\paragraph{Model Architecture}
We use the model from \citet{Dou_2022_CVPR}, but we simplify it with the removal of two of its components since we didn't observe a performance difference: the single-layer feedforward network before the feeding of the [CLS] representations to a task-specific head (e.g. ITM, MLM, classifier), and the image token type embeddings.

\paragraph{Pretraining}
We run pretraining for 7,400 steps, except when training on 1\%, 5\% and 10\% of CC, as more training results in an increase of the validation loss. We train for 1850 steps on 1\% and \%5 of CC, and 3700 steps for 10\% of CC.

The batch size is 3,840 and learning rates of $1e^{-4}$ and $5e^{-4}$ are used for the pretrained and randomly initialized weights respectively.
The learning rate is warmed up from zero during the first 10\% steps, and then linearly decays back to zero throughout the remaining steps.

We use 8 NVIDIA V100 GPUs, and training takes about 16 hours for 100\% of CC.

\paragraph{Finetuning}
For finetuning, we use a batch size of 96 for VQAv2 and GQA, and 48 for NLVR2.
We specify the learning rates for finetuning before and after bimodal pretraining in Tables \ref{table:lr_before} and \ref{table:lr_after} respectively.
The learning rate is warmed up from zero during the first 10\% steps, and then linearly decays back to zero throughout the remaining steps.

We use a single NVIDIA RTX 3090 GPU, and training takes 10, 15 and 4 hours for VQAv2, GQA and NLVR2 respectively.

\begin{table}[h]
\centering
\footnotesize
\begin{tabular}{llll}
\toprule
Weights                      & VQAv2 & GQA & NLVR2 \\
\midrule
Image encoder, Text encoder &  $2e-5$ &  $1e-5$   &  $1e-5$     \\
Cross-modal transformer      & $2e-4$      &  $1e-4$   &  $1e-4$     \\
Classifier head              &  $2e-4$     & $1e-4$    &   $1e-4$   \\
\bottomrule
\end{tabular}
\vspace{-0.3cm}
\caption{Learning rates per weights for finetuning \textit{before} bimodal pretraining for each downstream task.}
\label{table:lr_before}
\end{table}

\begin{table}[H]
\centering
\footnotesize
\begin{tabular}{llll}
\toprule
Weights                      & VQAv2 & GQA & NLVR2 \\
\midrule
Image encoder, Text encoder &  $2e-5$ &  $1e-5$   &  $1e-5$     \\
Cross-modal transformer      & $1e-4$      &  $1e-4$   &  $5e-5$     \\
Classifier head              &  $1e-3$     & $1e-4$    &   $5e-4$   \\
\bottomrule
\end{tabular}
\vspace{-0.3cm}
\caption{Learning rates per weights for finetuning \textit{after} bimodal pretraining for each downstream task.}
\label{table:lr_after}
\end{table}

\section{Results}\label{sec:appendix_tables}

\begin{table}[ht]
\centering
\footnotesize
\begin{tabular}{llll}
\toprule
CC Data &                 VQAv2 &                   GQA &                      NLVR2 \\
\midrule
    0\% &          69.5$\pm$0.1 &          53.6$\pm$0.1 &              random \\
    1\% &          69.2$\pm$0.1 &          53.8$\pm$0.6 &          67.3$\pm$0.5 \\
    5\% &          69.8$\pm$0.0 &          55.3$\pm$0.3 &          70.2$\pm$0.3 \\
   10\% &          70.1$\pm$0.1 &          55.5$\pm$0.2 &          71.2$\pm$0.4 \\
   25\% &          70.5$\pm$0.1 &          55.6$\pm$0.2 &          72.9$\pm$0.4 \\
   50\% &          70.6$\pm$0.1 & \textbf{56.1$\pm$0.4} &          73.8$\pm$0.4 \\
  100\% & \textbf{70.7$\pm$0.0} &          56.1$\pm$0.3 & \textbf{74.3$\pm$0.3} \\
\bottomrule
\end{tabular}
\vspace{-0.3cm}
\caption{Effect of the fraction of examples from CC on downstream task performance. Visualized with \cref{fig:cc-size}.}
\label{table:pretraining_results}
\end{table}

\begin{table}[ht]
\centering
\footnotesize
\begin{tabular}{l|lll|lll|lll}
\toprule
\multicolumn{1}{c|}{} & \multicolumn{3}{c|}{VQAv2} & \multicolumn{3}{c}{GQA} & \multicolumn{3}{c}{NLVR2} \\
\midrule
CC Data & 10\% & 25\% & 100\% & 10\% & 25\% & 100\% & 10\% & 25\% & 100\% \\
\midrule
   0\% &  54.4$\pm$0.0 &  62.1$\pm$0.6 &   69.5$\pm$0.1 & 45.6$\pm$0.5 & 48.0$\pm$0.1 & 53.6$\pm$0.1 & random & random &  random \\
   1\% &  55.8$\pm$0.2 &  60.8$\pm$0.1 &   69.2$\pm$0.1 & 45.2$\pm$0.3 & 48.1$\pm$0.5 & 53.8$\pm$0.6 & 52.5$\pm$0.9 &  54.6$\pm$0.6 &   67.3$\pm$0.5 \\
   5\% &  58.1$\pm$0.2 &  63.7$\pm$0.1 &   69.8$\pm$0.0 & 46.3$\pm$0.4 & 49.1$\pm$0.1 & 55.3$\pm$0.3 & 55.6$\pm$1.5 &  61.0$\pm$0.8 &   70.2$\pm$0.3 \\
   100\% &  62.4$\pm$0.2 &  66.0$\pm$0.1 &   70.7$\pm$0.0 & 48.4$\pm$0.6 & 51.8$\pm$0.3 & 56.1$\pm$0.3 & 63.4$\pm$0.5 &  67.6$\pm$0.6 &   74.3$\pm$0.3 \\
\bottomrule
\end{tabular}
\vspace{-0.3cm}
\caption{Effect of the fraction of examples from CC on downstream task performance when finetuning on less downstream data. Visualized with \cref{fig:low_resource_downstream}.}
\label{table:low_resource_downstream}
\end{table}

\begin{table*}[ht]
\centering
\footnotesize
\begin{tabular}{llll}
\toprule
  &                 VQAv2 &                   GQA &                      NLVR2 \\
\midrule
    Unimodally-pretrained &          69.5$\pm$0.1 &          53.6$\pm$0.1 &              random \\
    Bimodally-pretrained with ImageNet &          69.3$\pm$0.0 &          53.5$\pm$0.2 &          random \\
\bottomrule
\end{tabular}
\vspace{-0.3cm}
\caption{Performance on all downstream tasks, with and without ImageNet pretraining. No difference is observed.}
\label{table:imagenet_results}
\end{table*}

\begin{table}[ht]
\footnotesize
\begin{tabular}{lll}
\toprule
                  VQAv2 Data &                   GQA &                      NLVR2 \\
\midrule
       0\% &          53.6$\pm$0.1 &              random \\
                         5\% &          54.6$\pm$0.3 &          56.6$\pm$3.5 \\
                        10\% &          55.1$\pm$0.2 &          61.4$\pm$1.8 \\
                        25\% &          55.1$\pm$0.4 &          68.3$\pm$0.4 \\
                        50\% &          55.7$\pm$0.5 &          70.0$\pm$0.5 \\
                       100\% & \textbf{56.3$\pm$0.2} &          71.1$\pm$0.5 \\
\hline
Bimodally-pretrained with CC &          56.1$\pm$0.3 & \textbf{74.3$\pm$0.3} \\
\bottomrule
\end{tabular}
\vspace{-0.3cm}
\caption{Effect of the fraction of examples from VQAv2 on downstream task performance. Visualized with \cref{fig:transfer} (left).}
\label{table:vqa_pretraining_results}
\end{table}

\begin{table}[ht]
\centering
\footnotesize
\begin{tabular}{lll}
\toprule
                    GQA Data &                 VQAv2 &                      NLVR2 \\
\midrule
       0\% &          69.5$\pm$0.1 &              random \\
                         5\% &          69.1$\pm$0.1 &          52.3$\pm$1.5 \\
                        10\% &          69.3$\pm$0.1 &          53.3$\pm$2.3 \\
                        25\% &          69.2$\pm$0.1 &          55.5$\pm$3.2 \\
                        50\% &          69.3$\pm$0.1 &          59.5$\pm$2.4 \\
                       100\% &          69.4$\pm$0.1 &          63.1$\pm$1.0 \\
\hline
Bimodally-pretrained with CC & \textbf{70.7$\pm$0.0} & \textbf{74.3$\pm$0.3} \\
\bottomrule
\end{tabular}
\vspace{-0.3cm}
\caption{Effect of the fraction of examples from GQA on downstream task performance. Visualized with \cref{fig:transfer} (right).}
\label{table:gqa_pretraining_results}
\end{table}

\end{document}